\documentclass[lettersize,journal]{IEEEtran}
\usepackage{amsmath,amsfonts}
\usepackage{algorithmic}
\usepackage{algorithm}
\usepackage{array}

\usepackage[caption=false,font=normalsize,labelfont=sf,textfont=sf]{subfig}
\usepackage{textcomp}
\usepackage{stfloats} 
\usepackage{url}
\usepackage{verbatim}
\usepackage{graphicx}
\usepackage{cite}
\usepackage{nicematrix,tikz}
\usepackage{makecell}
\usepackage{hyperref}
\usepackage{placeins}
\DeclareCaptionLabelFormat{andtable}{#1~#2  \&  \tablename~\thetable}

\DeclareCaptionLabelFormat{twotable}{#1~#2  \&  \tablename~\the\numexpr\value{table}+1}
\NiceMatrixOptions
  {
    custom-line = 
     {
       letter = I , 
       command = boldhline , 
       total-width = 1 pt , 
       tikz = { line width = 1 pt } 
     }
  }

\begin{document} 


\title{Pedestrian intention prediction in Adverse Weather Conditions with Spiking Neural Networks and Dynamic Vision Sensors}

\author{
    Mustafa~Sakhai$^{\star}$,
    Szymon~Mazurek$^{\star}$,
    Jakub~Caputa,
    Jan K. Argasiński,
    Maciej~Wielgosz
    \IEEEcompsocitemizethanks{
        \IEEEcompsocthanksitem Mustafa Sakhai, Szymon Mazurek, Jakub Caputa, Maciej Wielgosz are with the Faculty of Computer Science, Electronics and Telecommunications, AGH University of Krakow, al. Adama Mickiewicza 30, 30-059 Krakow, Poland.
        \IEEEcompsocthanksitem Jan K. Argasiński is with the Faculty of Physics, Astronomy and Applied Computer Science, Jagiellonian University, ul. Łojasiewicza 11, 30-348 Krakow, Poland and Sano - Centre for Computational Personalised Medicine, ul. Czarnowiejska 36 (C5), 30-054 Krakow.
    }
}
      
\markboth{Submitted for review in IEEE Transactions on Intelligent Transportation Systems}%
{Shell \MakeLowercase{\textit{et al.}}: A Sample Article Using IEEEtran.cls for IEEE Journals}


\maketitle

\begin{abstract}

This study examines the effectiveness of Spiking Neural Networks (SNNs) paired with Dynamic Vision Sensors (DVS) to improve pedestrian detection in adverse weather, a significant challenge for autonomous vehicles. Utilizing the high temporal resolution and low latency of DVS, which excels in dynamic, low-light, and high-contrast environments, we assess the efficiency of SNNs compared to traditional Convolutional Neural Networks (CNNs).

Our experiments involved testing across diverse weather scenarios using a custom dataset from the CARLA simulator, mirroring real-world variability. SNN models, enhanced with Temporally Effective Batch Normalization, were trained and benchmarked against state-of-the-art CNNs to demonstrate superior accuracy and computational efficiency in complex conditions such as rain and fog.

The results indicate that SNNs, integrated with DVS, significantly reduce computational overhead and improve detection accuracy in challenging conditions compared to CNNs. This highlights the potential of DVS combined with bio-inspired SNN processing to enhance autonomous vehicle perception and decision-making systems, advancing intelligent transportation systems' safety features in varying operational environments.

Additionally, our research indicates that SNNs perform more efficiently in handling long perception windows and prediction tasks, rather than simple pedestrian detection.
\end{abstract}


\section{Introduction}
The field of Artificial Intelligence (AI) has evolved dramatically in recent years, impacting various scientific and technological domains. A prominent example is the autonomous vehicle industry, which relies heavily on AI and neural networks. Self-driving cars harness data from an array of internal systems and external sensors, driving forward a revolution in transportation. The vision of driverless cars, equipped with rapid computer reflexes and optimal urban navigation capabilities, exemplifies the potential of this technology. Despite the advancements, the widespread adoption of autonomous vehicles encounters challenges, including legal hurdles and the limitations of current data processing methods. Moreover, the expansion of onboard systems, such as electronic control units (ECU) and additional sensors, raises concerns over energy efficiency.

Waymo One, an early entrant in commercially available autonomous vehicles, operates on a blend of LIDAR (Light Detection and Ranging), radar, and long-range RGB cameras \cite{dilillo2023waymo}. While this autonomous vehicle functions without driver intervention, it is constrained to operate in controlled, predictable environments. The limitations of RGB cameras in varying lighting and weather conditions prompt the exploration of alternative technologies.

Dynamic Vision Sensors (DVS) emerge as a solution to these challenges. Unlike traditional RGB cameras, DVS cameras excel in low-light conditions and offer significant advantages in contrast detection. Their unique asynchronous operation mode allows them to focus on dynamic changes in the environment, resulting in lower latency and reduced data redundancy. These attributes not only enhance environmental perception but also contribute to greater energy efficiency \cite{Sharif2024dvsrev}.

The combination of DVS with Spiking Neural Networks (SNNs) presents a novel approach to processing visual information. SNNs, modeling the brain's impulse-based communication, are adept at handling the data stream from DVS cameras with minimal latency and energy consumption. This integration marks a significant advancement in developing sophisticated vision systems for autonomous vehicles, potentially revolutionizing the field of robotics and AI.

In this paper, we aim to explore the applicability of SNNs in pedestrian detection and intention prediction tasks. Due to the limited amount of accessible data, we use a simulation environment to create the dataset for further experiments. In the simulation, we emulate pedestrians in diverse urban and weather conditions, capturing the scenes using RGB and DVS cameras. We then evaluate the performance of SNN in detecting and predicting pedestrian behavior based on input sets of consecutive frames. We compare its performance against well-known deep learning models when working with data from varying obtained in good and bad weather conditions. We also perform comparative experiments using data from the JAAD \cite{rasouli2017JAAD1} dataset.
We show that SNN is able to perform on par or better than traditional neural networks in certain situations, especially when working with DVS data in difficult weather conditions. 
We openly share the dataset and the code used to conduct the experiments.

\section{Related work}
The success of deep learning in intelligent transportation systems has led to numerous solutions using Artificial Neural Networks (ANNs) for pedestrian detection and intention prediction \cite{Elladi2021AVsurvey, brunetti2018detectionSurvey, zhang2023pedestrianIntention}. Modern sensory devices, including DVS, have been integrated into autonomous vehicles, providing new data modalities for training deep learning models \cite{VoggingerSNNSRadars}. Wan et al. presented an innovative approach to convert DVS events into frames effectively and created a feature extraction network that can recycle features to minimize computational workload \cite{Wan2021DVSDetection}. Their method was evaluated on a custom dataset containing multiple real-world pedestrian scenes, resulting in an enhanced pedestrian detection accuracy improvement of 5.6–10.8\%. Additionally, the detection speed was boosted by nearly 20\% compared to earlier methods, achieving a processing speed of around 26 frames per second with a precision of 87.43\%. Studies have also shown that utilizing transfer learning allows for training networks that operate on DVS data with low latency. Chen has shown that detection in a real environment can be performed with a remarkable 100 frames per second, with an average test precision of 40.3\% \cite{Chen2018PseudoLabels}.



In comparative model performance, Zhao et al. trained models for both DA-ANN and basic ANN frameworks. The results showed that DA-ANN model 1 achieved 96.6\% accuracy versus 89.7\% for basic ANN model 1, and DA-ANN model 2 scored 94.3\% accuracy compared to 88.5\% for its basic ANN counterpart \cite{Zhao2019Trajectory}.



Recent advances in Spiking Neural Networks (SNNs) have shown their growing potential in autonomous driving due to optimization techniques for training these networks \cite{NeftciSurrogateGL, Wang2022SNNSTaxonomy}. Pascarella et al. demonstrated that combining frame-based and DVS cameras allows training Convolutional Neural Networks (CNNs) to effectively solve steering angle estimation problems \cite{pascarella2023dvsann}. Cordone et al. adapted popular ANN models into spiking versions for pedestrian and car detection in autonomous vehicle systems using DVS data, showing promising results \cite{CordoneSNNDetection}. Kim et al. proposed an SNN version of the YOLO algorithm, nearly matching the performance of ANN-based models while being faster to train and more energy-efficient \cite{kim2019spikingyolo, Jiang2022YOLORev}. Evidence also supports the effectiveness of SNN models on neuromorphic accelerators. Massa et al. used the Loihi chip \cite{Davies2018Loihi} for gesture recognition with DVS data, achieving similar performance to ANNs with significant energy savings \cite{MassaDVSLoihi}. Viale et al. created a similar solution for object classification in autonomous vehicles, also deployed on the Loihi chip, confirming these findings \cite{Viale2021CarSNN}.

\section{Background}
This section provides details about the mathematical foundations for tested SNN models, as well as CARLA\cite{Dosovitskiy2017Carla} generation environment used for data generation and the principles governing DVS. 
\subsection{Spiking Neural Networks and neuron models}
Due to the large success of ANNs, adoption of the successful architectures when training SNNs seems natural. Indeed, such solutions have shown high effectiveness, allowing the performances to reach levels competitive with ANNs \cite{FriedemannRobustnessSurrogate}. However, such architectural transfers cannot be done straightforwardly, as the principle governing SNNs requires certain modifications to the models \cite{Ding2021ANNSNN}. Here, we will first describe the general principles of spiking neurons, followed by the details of the one used in this study. Then, we will proceed to describe the learning method and details of the adopted ResNet \cite{he2016ResNet} architecture along with modifications that were introduced to further increase its effectiveness in processing spike trains.
\subsubsection{Basic principles of spiking neurons}
In SNNs, the neurons are modeled as entities that pass the information using spike trains, represented as vectors of binary values. Conversion of ANNs into SNNs thus requires replacing the nonlinear activation function with a model of spiking neurons. The charge of the neuron in a given timestep $t$ is given by:
\begin{equation}
    H[t] = f(V[t-1], X[t]),
\end{equation}
where $X$ is the input vector, $V$ is the discharge function and $f$ is the neuron function, which depends on the type of chosen model and will be described later. 
The discharge function describes the neuron's behavior after emitting a spike. It can incorporate hard or soft reset, in both cases resulting in an instant decrease of the membrane potential. In this paper, we use the hard reset approach, thus we derive only this equation:
\begin{equation}
    V[t] = H[t] \cdot (1-S[t]) + V_{reset} \cdot S[t],
\end{equation}
where $S$ is the neuronal firing function and $V_{reset}$ is the reset voltage value, to which the membrane potential comes back after emitting the spike.
The equation describing the firing can be denoted as:
\begin{equation}
    S[t] = \Theta(H[t] - V_{th}),
\end{equation}
where $V_{th}$ is the threshold voltage value and $\Theta$ is a Heaviside function, denoted as:
\begin{gather}
   \Theta(x) =
   \begin{cases} 1, \;  x\geq0  \\0, \;  x<0 \end{cases}
\end{gather}

\subsubsection{Neuron model}
There exist numerous biologically plausible models of neurons \cite{Izhkievich2004Neurons}. These models differ mostly in their biological plausibility and computational demands. In this work, we used the Parametric Leaky Integrate and Fire (PLIF) model \cite{fang2021plif}, an extension of the commonly used Leaky Integrate and Fire (LIF) model. Compared to LIF, PLIF allows for learning a $\tau$ parameter, a membrane time constant controlling the rate of membrane potential decay over time.
PLIF neuron is expressed as:
\begin{equation}
    f(V[t-1], X[t]) = V[t-1] + \frac{1}{\tau}(X[t] - (V[t-1] -V_{reset})),
\end{equation}
where $\frac{1}{\tau}=sigmoid(a)$. Here, $a$ is a learnable parameter shared across all neurons in a given layer. The sigmoid function is introduced to ensure that $\tau >1$.
\subsubsection{Surrogate gradient training of spiking neural networks}
The sparseness of SNNs, along with the benefits, results in challenges in training such networks. Non-continuous functions cannot be differentiated, therefore well-known gradient optimization used to train ANNs is unusable directly. The surrogate gradient method is one of the techniques for approximating the function representing the SNN as continuous, therefore enabling backpropagation and gradient optimization \cite{NeftciSurrogateGL}.

During the forward pass, the response of a neuron remains unchanged, being expressed as a previously described Heaviside function. The derivative of this function is equal to Dirac's delta:
\begin{gather}
   \Theta'(x) =
   \begin{cases} \infty, \;  x=0  \\0, \;  x\neq0 \end{cases}
\end{gather}
which prevents the direct application of backpropagation.
To amend this, during the backward pass, the Heaviside function is approximated as a chosen continuous function. In the case of this study, we used a sigmoid function approximation, defined as:
\begin{equation}
    \sigma(x,\alpha) = \frac{1}{1+\exp(-\alpha x)},
\end{equation} with $\alpha$ being the hyperparameter controlling the smoothness.
Its derivative can now be expressed as:
\begin{equation}
    \sigma'(x, \alpha) = \sigma(x,\alpha)(1-\sigma(x,\alpha))
\end{equation}
which is continuous and differentiable. Thus, the firing function during the backward pass becomes:
\begin{equation}
    S[t] = \sigma(H[t]-V_{th}),
\end{equation}
allowing for computation of the gradient and error backpropagation. In this work, the PLIF neuron was adopted with the following hyperparameters: initial $\tau=2$, $V_{th}=1$, and smoothing factor for a surrogate function $\alpha=4$.
\subsubsection{ResNet architecture adaptations}
In this work, we decided to adopt ResNet architecture to its spiking variant due to the model's simplicity and large success in the ANN domain in various tasks. 
In \cite{fangSEWResnet}, Wang et.al. observed that ResNet architecture converted into a spiking variant suffers from problems of vanishing gradients and fails to properly implement identity mapping in the residual block for most of the neuron models.
We adapt the solution presented in the aforementioned work by replacing the addition operation between the output of the residual block $A[t]$ and the input to it $I[t]$ with one of the proposed operands $G$:
\begin{equation}
    G[t]=(\neg A[t]) \wedge I[t]
\end{equation}

As another modification, we use a Temporally Effective Batch Normalization (TEBN) layer in replacement of the standard batch normalization (BN) layer. This step was guided by findings of \cite{duanTEBN}, where authors prove the superiority of TEBN over other normalization techniques \cite{kim2021BNTT, ioffe2015batchnorm} in SNNs due to its ability to capture richer properties of the spike trains.
The normalized output $\hat{X}[t]$ from TEBN layer is defined as:
\begin{equation}
    \hat{X}[t] = \hat{\gamma}[t]\frac{X[t]-\mu_{total}}{\sqrt{\sigma^2_{total}+\epsilon}} + \hat{\beta}[t],
    \label{eq-tebn}
\end{equation}
\begin{equation}
     \hat{\gamma}[t] = \gamma \times p[t], 
    \hat{\beta}[t] = \beta \times p[t].
\end{equation}
Here, in each TEBN layer, $\gamma$ and $\beta$ are time-invariant BN parameters, and $p[t]$ is a set of learnable weight parameters. The mean $\mu$ and variance $\sigma^2$ are calculated from samples across all time-steps, and $\epsilon$ is a small constant that ensures numerical stability.
\subsubsection{Network readout}
As the spiking network produces T output spikes given T input ones, we averaged the output spike train to produce the classification logit. Therefore, the network's output $\hat{Y}$ was equal to:
\begin{equation}
    \hat{Y} = \frac{1}{T}\times \sum_{i=0}^{T}y[t],
    \label{eq-logit}
\end{equation}
where $y[t]$ is the output spike train value at timestep $t$.

\subsection{CARLA Simulator and Perception System}

CARLA is an open-source simulator for autonomous driving research, developed by the Computer Vision Centre and the Embodied AI Foundation \cite{Dosovitskiy2017Carla}. Built on Unreal Engine 4, it offers a realistic urban environment with various scenarios and weather conditions, enabling safe and controlled testing of algorithms. CARLA's extensive customization options, including modifiable environments, vehicle models, and sensors, make it a powerful tool for advancing autonomous driving research. It allows for the integration of custom algorithms and large-scale experiments, making it widely adopted by research institutions and companies globally.

Importantly, this environment allows for detailed customization of weather conditions. It provides a large number of configurable and independent parameters, offering flexibility in creating specific environmental scenarios. Key parameters include cloudiness, precipitation, wind intensity, sun position, fog density, and wetness. These settings range from clear skies to heavy rain, strong winds, and dense fog, providing a versatile platform for testing autonomous vehicle systems under varied conditions. For example, cloudiness and precipitation levels can be adjusted to simulate everything from a clear day to a severe storm, while sun azimuth and altitude angles control the sun's position in the sky. Fog parameters dictate the density and distance of fog, enhancing the realism of low-visibility scenarios. Additionally, CARLA supports the creation of puddles and varying wet road conditions to mimic post-rain environments. We leverage these parameters to obtain diverse and challenging driving conditions scenes, allowing for a thorough evaluation of the performance and robustness of autonomous driving algorithms.

ScenarioRunner is another tool offered by the CARLA simulator for defining and executing traffic scenarios. It provides the ability to create and validate complex traffic scenarios that can be used to evaluate and benchmark autonomous driving agents. ScenarioRunner allows the selection of maps, weather, sensors, and textures and manages them in a controlled way. 

\subsection{Dynamic Vision Sensor (DVS)}

The Dynamic Vision Sensor (DVS), also known as an Event Camera, operates differently from traditional cameras by capturing changes in intensity asynchronously as a stream of events. Each event corresponds to a change in brightness and encodes information about its pixel location, timestamp, and polarity. Event cameras offer several advantages over conventional cameras, including high dynamic range, no motion blur, and high temporal resolution in the order of microseconds \cite{Sharif2024dvsrev}.

To generate an event, the change in logarithmic intensity must exceed a predefined threshold, resulting in a polarity of either positive or negative. 

CARLA offers access to the DVS camera during the simulations. It operates in a uniform sampling manner between two consecutive synchronous frames, requiring high frequency to emulate the high temporal resolution of a real event camera. 


\section{Dataset generation}
Our main objective was to investigate the problem of detecting and predicting pedestrian crossing behavior in difficult weather with the use of neural networks on data coming from different sensors. Initially, we sought a dataset capturing pedestrians crossing streets in various weather conditions with both DVS and RGB images. While there are many road-themed datasets available online, such as n-cars \cite{CordoneSNNDetection}, JAAD \cite{rasouli2017JAAD}, and DSEC \cite{Gehrig2021DESCDataset}, as well as numerous articles analyzing pedestrian crossing intentions \cite{Qui2023Decision, zhang2023pedestrianIntention}, none fully met our requirements. These datasets either lack DVS data, provide limited instances of pedestrians crossing the streets or appropriate labeling of such events, or do not contain enough video materials with difficult weather conditions. Consequently, we decided to generate a custom dataset using a simulation environment.

The simulation was created using the previously described CARLA software and the scenario repository from the ARCANE project, which focuses on adversarial cases for autonomous vehicles \cite{Riaz2023ARCANE}. This module allows for the specification of various recording parameters, including sample quantity, recording duration, camera angle, weather conditions, maps, resolution, and pedestrian attributes such as appearance and gender. 

During the simulation of these environments, the data is recorded in real-time, with an automatic filtering process extracting instances of pedestrians. The RGB camera frames are saved as JPG files, while DVS data is extracted from the resulting event trains and saved as PNG files. For both modalities, we sample frames with a rate of 30 per second. After the simulation, each frame is labeled manually. If a pedestrian is crossing the street in a given frame, it is labeled as positive, with a negative label being assigned otherwise.

We created two distinct subsets: the "Default" subset, featuring clips captured under sunny weather conditions, and the "Weather" subset, which includes various weather effects such as rain, storm, and fog that can hinder visibility. Each recording comprises separate DVS and RGB clips, totaling 900 frames per clip. The resolution of the RGB images is 1600x600, while the DVS images have a resolution of 1542x587. The "Default" subset consists of 117 videos, capturing pedestrians crossing roadways under sunny conditions. The "Weather" subset contains 81 videos, introducing challenging weather conditions.

\begin{figure*}
    \centering
        \includegraphics[scale=0.8]{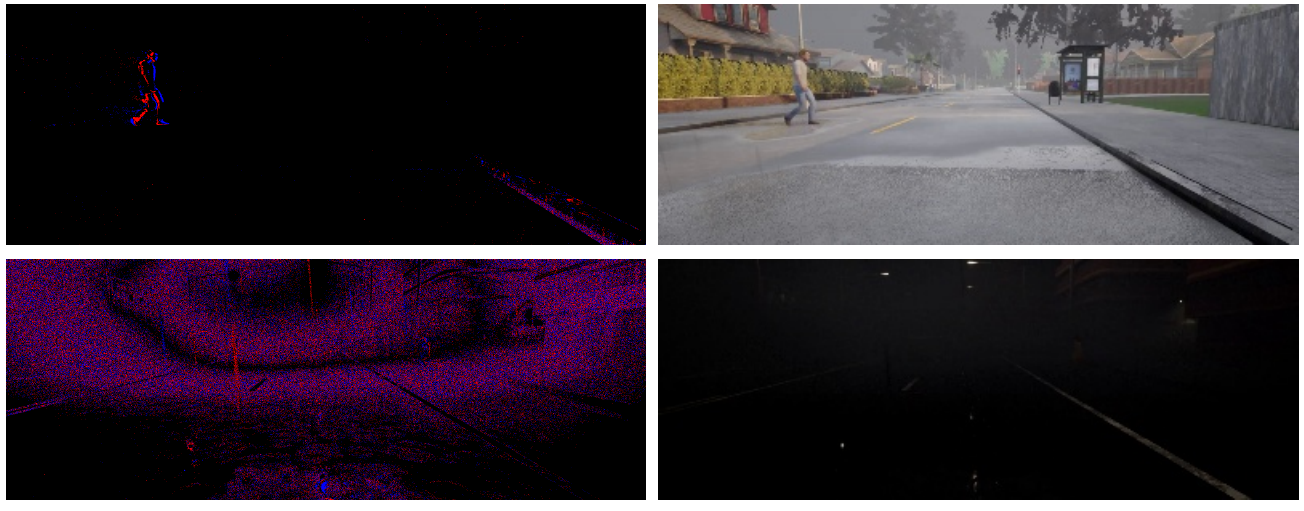}
    \caption{Example frames from the dataset. The first row shows samples from the good weather subset, and the second row shows samples from the bad weather subset. The first column displays images captured by the DVS camera, while the second column shows the corresponding RGB images. Each DVS-RGB pair represents the same frame within a given subset. A pedestrian is present in the scene in both of the presented frames. Note that in the bad weather frames, the pedestrian is barely visible to the human observer in the RGB images on the right.}
    \label{fig:example-frames}
\end{figure*}


\section{Experimental Setup for Network Training}

In this section, we outline the setup for our experiments using the generated datasets to address various tasks. We evaluated three networks: ResNet18, its spiking adaptation Spiking Sew ResNet18 with TEBN (SPS R18T), and SlowFast R50 \cite{feichtenhofer2019slowfast}, a model designed specifically for video classification. Pre-trained weights were used for the ANNs: ResNet18 was trained on the ImageNet1k dataset \cite{deng2009imagenet}, while SlowFast R50 utilized the Kinetics400 dataset \cite{KayK2017inetics}.

Videos were split into training, validation, and testing subsets, with 15\% of the total videos allocated for testing. From the remaining videos, 15\% were used for validation and the rest for training. Frames were processed based on the specific task (see task descriptions below).

All networks were optimized using the AdamW optimizer, aiming to minimize weighted binary cross-entropy loss. The initial learning rate was set to $10^{-3}$, with a weight decay factor of $10^{-1}$. Batch sizes ranged from 4 to 64, depending on hardware capacity. The training was run for a maximum of 100 epochs, employing an early stopping protocol if validation loss did not improve for 8 consecutive epochs. The best-performing model, based on validation loss, was used for testing. Performance was measured using AUROC and F-score, suitable for imbalanced classification problems.

For all experiments, the code was developed in Python 3.10 with PyTorch 2.2.0 \cite{PaszkeTorch} and Lightning 2.1.3 framework, along with SpikingJelly 0.0.0.15 \cite{Fang2023SpikingJelly}, a library implementing abstractions related to SNNs. All the parameters were given in Tab. \ref{table:training_parameters}. Experimental code is available at 
\href{https://github.com/szmazurek/snn\_dvs}{https://github.com/szmazurek/snn\_dvs}

\begin{table*}[h!]
\centering
\caption{Training Parameters}
\begin{tabular}{|c|c|c|}
\hline
\textbf{Parameter} & \textbf{Value} & \textbf{Comments} \\ \hline
Initial Learning Rate & $10^{-3}$ & Learning rate used at the start of training \\ \hline
Weight Decay Factor & $10^{-1}$ & Regularization parameter to prevent overfitting \\ \hline
Batch Size & 4 to 64 & Varies based on hardware capacity \\ \hline
Maximum Epochs & 100 & Maximum number of training epochs \\ \hline
Early Stopping & 8 epochs & Stops training if no improvement in validation loss for 8 consecutive epochs \\ \hline
Loss Function & Weighted binary cross-entropy & Weights calculated based on the negative-to-positive sample ratio \\ \hline
Optimization Algorithm & AdamW & Optimizer used for training \\ \hline
Performance Metrics & AUROC, F-score & Metrics used for evaluating model performance \\ \hline
\end{tabular}
\label{table:training_parameters}
\end{table*}

\subsection{Tasks and Data Processing}
This section outlines the specific tasks and data processing steps undertaken. Each task utilized the resized frames and adhered to previously established training, validation, and testing protocols.

\subsubsection{Clip Classification}
At first, we explore the problem of \textit{pedestrian detection}, where we aim to identify if the pedestrian is crossing the street at a given moment.
We explored the performance of the networks in this task using sets of consecutive frames as inputs to the network. For this purpose, the ANN version of ResNet was trained using a ``pseudotemporal'' scheme, where each frame in the input clip contributes to a prediction that is then averaged to produce a single logit, as described by Equation \ref{eq-logit}. Additionally, we assessed the SPS R18T and SlowFast R50 architectures, both of which inherently support temporal information processing. These networks were tested under two scenarios: one with clip lengths of 9 frames and 8 frames overlapping, and another with clip lengths of 30 frames and 29 overlapping. For labeling, clips were categorized as positive if at least one frame showed a pedestrian crossing the street; otherwise, they were labeled as negative. In this case, class imbalances were observed, with more negative samples present. Therefore, the loss function was weighted for positive samples proportionally to the imbalance observed in a given setting.

\subsubsection{Clip Prediction}
Building on the detection tasks, we also evaluated the predictive capabilities of our networks through clip classification. In this case, we specify the prediction horizon of length $H$, which defines the number of frames labeled as positive that directly precede the first frame in which the pedestrian starts crossing the street in a given video. The negative frames were extracted from the videos in which the pedestrian did not cross the street. The number of negative samples to extract was configured to match the number of positive samples, preventing class imbalance.
Due to the limited number of samples, particularly for shorter prediction horizons (\( H=1s \)), we confined our tests to clips of length 9 with 8 frames overlapping. The rest of the experiment organization was configured in the same way as for clip classification.

\subsection{Benchmarking the Solution Against the JAAD Dataset}
To contextualize our findings within existing research, we adapted the JAAD dataset. This dataset distinguishes between good and bad weather conditions and includes RGB clips of various lengths. We aligned our labeling approach with the existing annotations to determine if a pedestrian is crossing the street in each frame. Given the limited number and duration of clips, especially under bad weather conditions, we were not able to create positive samples in the same way as in the previous prediction experiments. Due to this fact, we conducted the experiments only for the detection task. The experimental setup for the JAAD dataset was the same as for the simulated dataset described earlier.

\subsection{Measuring energy usage}
As energy-efficient processing is one of the prominent benefits of SNNs, we compare the energy usage of each trained network. We follow the methodology in the research by Chen et. al \cite{chen2023energyops}, measuring the number of multiply-and-accumulate (MAC) and accumulation (AC) operations in a given network and translating that to the energy usage of such operations in $45nm$ technology \cite{chen2023energyops,Horwitz2014energycomputing}. In standard feedforward ANNs used in this research, the number of operations is constant in every forward pass. Therefore, for this type of network, we compute the number of operations during a single forward pass with a dummy input tensor of shape identical to one of the processed clips in experimental tasks. For SNNs, the number of emitted spikes varies depending on the input sample. Due to this fact, for SNNs, we estimated the consumption by performing the inference on every sample of the test dataset with the corresponding trained model and averaged the number of operations.
We chose bad weather DVS data as the evaluation one. Corresponding evaluations for ANNs were also done using single-channel dummy data.
\section{Results}
\subsection{Street Crossing Detection in Clips of Consecutive Frames}

The results of the evaluation of the detection task are summarized in Table \ref{tab:clip-classification}. 

For shorter time windows of 9 frames, PT ResNet18 exhibited superior performance across most cases. Specifically, in the bad weather subset, PT ResNet18 achieved outstanding results with an AUROC of 0.9882 and an F-score of 93.05 for DVS data, and an AUROC of 0.9194 with an F-score of 58.02 for RGB data. The SPS R18T model showed competitive performance in the DVS modality, with an AUROC of 0.9504 and an F-score of 57.94, though it lagged significantly in RGB performance.

In normal weather conditions, all networks demonstrated robust detection capabilities, particularly in the DVS modality, where AUROC scores exceeded 0.94 for each network. The PT ResNet18 continued to perform well with RGB data, obtaining an AUROC of 0.9516 and an F-score of 76.95. Conversely, the SPS R18T experienced a noticeable drop in performance in RGB data, registering an AUROC of 0.7995 and an F-score of 49.26.

When extending the clip length to 30 frames, the observed trends were consistent with shorter clips. PT ResNet18 continued to exhibit top performance, slightly surpassed by SlowFast R50 in the good weather subset. For instance, in bad weather conditions with RGB data, SlowFast R50 markedly improved, increasing its AUROC from 0.5126 in shorter clips to 0.9546 in longer clips, and similarly for DVS data in good weather conditions, achieving an AUROC of 0.9594 and an F-score of 81.68. The SPS R18T model demonstrated comparable performance across various conditions and modalities, regardless of clip length.

\begin{table*}[]
\centering
\caption{Performance of the networks across various datasets, evaluated by AUROC and F-score for clip classification, with F-scores expressed as percentages. Bold figures highlight the top results within each data subset and modality. "PT ResNet18" refers to the pseudotemporal variant of the standard ResNet18, and "SPS R18T" represents the Spiking Sew ResNet18 enhanced with Temporally Effective Batch Normalization. The best outcomes for each subset and modality are emphasized in bold.}
\begin{NiceTabular}{|p{1.7cm}|p{1.cm}|p{1.cm}|p{1.cm}|p{1.cm}|p{1.cm}|p{1.cm}|p{1.cm}|p{1.cm}|}
    \hline
    \Block[draw]{2-1}{\diagbox{Network}{Subset}}  & \Block{1-4}{\makecell{Bad weather}} & & & & \Block{1-4}{\makecell{Normal weather}} \\
    \cline{2-9}
    & \Block{1-2}{\makecell{DVS}} &  & \Block{1-2}{\makecell{RGB}} &  & \Block{1-2}{\makecell{DVS}} &  & \Block{1-2}{\makecell{RGB}} &  \\
    \hline
     \diagbox{}{} & AUROC & F-score & AUROC & F-score & AUROC & F-score  & AUROC & F-score  \\
     \hline
    \Block{1-9}{\makecell{Clip length: 9 frames}} & & &  \\
    \hline
    PT ResNet18  & \textbf{0.9882} & \textbf{93.05} & \textbf{0.9194} & \textbf{58.02} & \textbf{0.9685} & \textbf{83.83} & 0.9516 & 76.95 \\
    \hline 
    SlowFast R50  & 0.7446 & 42.17 & 0.5126 & 20.01 & 0.9487 & 72.61 & \textbf{0.9932} & \textbf{92.81} \\
    \hline 
    SPS R18T & 0.9504 & 57.94 & 0.6771 & 28.11 & 0.9421 & 74.4 & 0.7995 & 49.26 \\
    \hline
    \Block{1-9}{\makecell{Clip length: 30 frames}} & & &    \\
    \hline
    PT ResNet18  & \textbf{0.975} & \textbf{91.94} & \textbf{0.9825} & \textbf{83.56} & 0.9548 & 79.87 & 0.9616 & 80.5 \\
    \hline 
    SlowFast R50  & 0.9535 & 80.31 & 0.9546 & 84.3 & \textbf{0.9594} & \textbf{81.68} & \textbf{0.9784} & \textbf{87.68} \\
    \hline 
    SPS R18T & 0.9542 & 75.5 & 0.6811 & 45.05 & 0.9289 & 73.43 & 0.8012 & 54.39 \\
    \hline
\end{NiceTabular}

\label{tab:clip-classification}

\end{table*}

\subsection{Street crossing prediction in clips of consecutive frames}

The results for prediction task experiments are shown in the Tab. \ref{tab:clip-prediction}.

For the 1-second predictive horizon, we can see the remarkable performance of the SPS R18T in the bad weather subset for both modalities, where it outperforms both SlowFast R50 and PT ResNet18. The performance advantage is most visible with the DVS modality. Notably, for the same data modality, SlowFast R50 failed to converge, reaching AUROC of only 0.5827 and 0\% F-score. This changed in the good weather subset, where classic ANNs performed better than the SNN. The best results were achieved on the DVS modality, with the top one reached by PT ResNet18 with an AUROC of 0.9455 and an F-score of 93.17\%.

For the 5-second predictive horizon, again the best class separability in the bad weather subset was reached by the SPS R18T. Contrary to the shorter lookback observations, this time RGB data turned out to be more informative for the SNN, where it reached an AUROC of 0.882 and an F-score of 79.39. 
In the normal weather subset, the networks maintain robust performance on the DVS modality, except for the SPS R18T, which notes a large drop in performance compared to the shorter predictive horizon version. This is similar to what can be seen in the previous experiments with single-frame prediction for this network. A large performance decrease can be seen for the RGB modality, where none of the networks reached results better than random guesses with AUROC scores below 0.5.

\begin{table*}[]
\centering
\caption{Performance of the compared networks on different datasets as measured by test AUROC and F-score for classification if the given set of consecutive frames comes directly from the period before the pedestrian crosses the street or not. F-score is denoted in percentages. All clip lengths were set to 9 frames, with 8 frames overlap. PT ResNet18 stands for the "pseudotemporal" version of the normal ResNet18, and SPS R18T denotes Spiking Sew ResNet18 with Temporally Effective Batch Normalization. Bold numbers indicate the best results for the given data subset and modality.}
\begin{NiceTabular}{|p{1.7cm}|p{1.cm}|p{1.cm}|p{1.cm}|p{1.cm}|p{1.cm}|p{1.cm}|p{1.cm}|p{1.cm}|}
    \hline
   \Block[draw]{2-1}{\diagbox{Network}{Subset}} &   \Block{1-4}{\makecell{Bad weather}} & & & & \Block{1-4}{\makecell{Normal weather}}  \\
   \cline{2-9}
   & \Block{1-2}{\makecell{DVS}} &  & \Block{1-2}{\makecell{RGB}} &  & \Block{1-2}{\makecell{DVS}} &  & \Block{1-2}{\makecell{RGB}} &  \\
   \hline
     \diagbox{}{} & AUROC & F-score & AUROC & F-score & AUROC & F-score  & AUROC & F-score  \\
     \hline
     \Block{1-9}{\makecell{Horizon: 30 frames (1s)}} & & &    \\
     \hline
     PT ResNet18  & 0.7959 & 80 & 0.6476 & 70.97 & \textbf{0.9455} & \textbf{93.17} & 0.7177 & 65.45  \\
    \hline
    SlowFast R50 & 0.5827 & 0 & 0.8333 & \textbf{89.08} & 0.9422 & 88.37 & \textbf{0.8173} & \textbf{76.34} \\
    \hline
    SPS R18T & \textbf{0.852} & \textbf{83.27} & \textbf{0.8643} & 68.75 & 0.8802 & 82.55 & 0.3449 & 52.33  \\
    \hline
   \Block{1-9}{\makecell{Horizon: 150 frames (5s)}} & & &    \\
     \hline 
    PT ResNet18 & 0.6737 & 59.86 & 0.8633 & \textbf{81.7} & \textbf{0.8608} & \textbf{79.56} & 0.4053 & 56.07 \\
    \hline
    SlowFast R50 & 0.7903 & \textbf{69.97} & 0.6271 & 64.81 & 0.8291 & 73.71 & \textbf{0.454} & 57.08 \\
    \hline
    SPS R18T & \textbf{0.8249} & 67.99 & \textbf{0.882} & 79.39 & 0.4423 & 56.42 & 0.4128 & \textbf{62.96} \\
     \hline
\end{NiceTabular}
\label{tab:clip-prediction}
\end{table*}

\subsection{Detection and prediction tasks on JAAD dataset}


Overall, the performances of nearly all models are lower than the ones observed on the simulation dataset. It can be seen that SPS R18 underperforms in nearly every scenario. This could, however, be expected, as this dataset contained only RGB data on which ANNs have generally shown better performance. In the single frame detection scenario, the spiking network performs better on the bad weather data, although the performance is still close to random choice with an AUROC of 0.548. In good weather, standard Resnet was better, reaching an AUROC of 0.6964 and an F-score of 66.54\%. 

When the detection was performed on the sets of consecutive frames, SPS R18T was better than ANN counterparts for the shorter clips of 9 frames, with AUROC of 0.5966 and F-score of 74.03\%. For longer clips in the same subset, the gap is reduced, yet still, spiking networks have shown the best class separability with an AUROC of 0.642.
These trends change in the good weather subset. The performance of SNN remains similar to the one on the bad weather data, while PT Resnet and SlowFast R50 show significant performance improvements on both short and long samples.

In the prediction task, standard Resnet has shown generally better performance than spiking one. They perform on a similar level only in bad weather settings with longer prediction horizons. In the good weather subset, Resnet consistently performs better with AUROC above 0.8 and F-score close to 80\%. When performing the predictions on clips of frames, the data deficiencies become more apparent, as all networks show identical F-scores on the bad weather subset. 
With the good weather data, ANNs perform better on the short predictive horizon, with PT Resnet reaching 0.8066 AUROC and an F-score of 57.14\%.
This network remained the best-performing one when the predictive horizon increased, although it was closely followed by the spiking counterpart.

\begin{table}[htbp]
\centering

\caption{Performance of the compared networks on JAAD dataset as measured by test AUROC and F-score for single or repeated frame classification. F-score is denoted in percentages. Left table: performance in single-frame classification scenario. Right table: performance in clip classification scenario. PT ResNet18 stands for the "pseudotemporal" version of the normal ResNet18, and SPS R18T denotes Spiking Sew ResNet18 with Temporally Effective Batch Normalization. Bold numbers indicate the best results for the given data subset and modality.}
\begin{NiceTabular}{|p{1.7cm}|p{1.cm}|p{1.cm}|p{1.cm}|p{1.cm}|}
    \hline
   \Block[draw]{2-1}{\diagbox{Network}{Subset}} &   \Block{1-2}{\makecell{Bad weather}} &  & \Block{1-2}{\makecell{Good weather}}  \\
   \cline{2-5}

      & AUROC & F-score & AUROC & F-score \\
     \hline
     \Block{1-5}{\makecell{Clip length: 9 frames}} & & &    \\
     \hline
     PT ResNet18 & 0.4366 & 56.3 & \textbf{0.7597} & 71.92 \\
    \hline
    SlowFast R50  & 0.3585 & 52.89 & 0.7513 & \textbf{76.97} \\
    \hline 
    SPS R18T & \textbf{0.5966} & \textbf{74.03} & 0.6169 & 70.2 \\
    \hline
   \Block{1-5}{\makecell{Clip length: 30 frames}} & & &    \\
     \hline 
    PT ResNet18  & 0.5966 & \textbf{74.42} & \textbf{0.7861} & 46.24 \\
    \hline
    SlowFast R50  & 0.4773 & 56.25 & 0.5269 & \textbf{80.7} \\
    \hline 
    SPS R18T & \textbf{0.642} & 68.57 & 0.6093 & 65.78 \\
     \hline
\end{NiceTabular}

\label{tab:clip-prediction-jaad}
\end{table}

\subsection{Energy usage measurement}

The results of energy usage evaluations for the networks used in the previous experiments are shown in Tab. \ref{tab:energy-usage}. SPS R18T has shown the lowest energy consumption among all networks. For input clips of 9 frames, its average energy consumption was 50,45 mJ of energy compared to nearly 10 times more used by SlowFast R50 and more than 3 times more by PT Resnet18. 

Interestingly, SPS R18T has shown decreased energy usage when the number of samples in the clip increased, using only 30,37 mJ of energy. The energy usage of ANNs rose proportionally to the number of samples in the input clip.
\begin{table}
\centering
\caption{Comparison of number of parameters, energy usage, and number of specific operations in the used networks.}
\begin{NiceTabular}{|p{1.6cm}|p{1.1cm}|p{1.cm}|p{1.1cm}|p{1.3cm}|}
    \hline
   Model name & Parameters & ACs & MACs & Energy \newline usage [mJ]  \\
     \hline
     \Block{1-5}{\makecell{Clip length: 9 frames}} & & &    \\
     \hline
     PT ResNet18 & 11.17 M & 0 & 36.16 G & 166.33 \\
    \hline
    SlowFast R50  & 31.53 M & 0 & 109.94 G & 505.73 \\
    \hline 
    SPS R18T & 11.17 M & 22.75 G & 6.52 G & \textbf{50.45} \\
    \hline
   \Block{1-5}{\makecell{Clip length: 30 frames}} & & &    \\
     \hline 
    PT ResNet18  & 11.17 M & 0 & 120.53 G & 554.42 \\
    \hline
    SlowFast R50  & 31.63 M & 0 & 363.65 G & 1672.8 \\
    \hline 
    SPS R18T & 11.17 M & 4.63 G & 6.13 G & \textbf{32.37} \\
     \hline
\end{NiceTabular}

\label{tab:energy-usage}
\end{table}

\section{Discussion}

Our experimental results reveal nuanced insights into the performance of traditional and spiking neural networks across various weather conditions and modalities. Notably, the Spiking Sew ResNet18 with Temporally Effective Batch Normalization (SPS R18T) demonstrated exemplary performance when utilizing DVS data in adverse weather conditions. Achieving an AUROC of 0.9542 and an F-score of 75.5\% in the 30-frame clip scenario, SPS R18T showcased its effectiveness in detecting dynamic events such as pedestrian movement under challenging visual scenarios, including low light and precipitation (see Table \ref{tab:clip-classification}).

Conversely, traditional ANNs like PT ResNet18 and SlowFast R50 demonstrated robust performance across all conditions, particularly excelling in normal weather settings. For instance, SlowFast R50, when given RGB data in good weather conditions, improved its performance with AUROC soaring to 0.9784 and an F-score of 87.68 \% for longer clips, highlighting its efficacy in handling high-resolution, color-rich data (see Table \ref{tab:clip-classification}). This capability is crucial for accurate pedestrian detection, where visual details are paramount and not obscured by environmental factors.

While the SPS R18T performs competitively in DVS modalities, it exhibits a noticeable performance drop in RGB data across most weather scenarios. For example, its performance in RGB dropped significantly in both bad and normal weather conditions, as noted in its lower AUROC and F-score results compared to other networks. This suggests that while SNNs are particularly adept at processing dynamic visual changes enabled by DVS technology, they may not yet fully exploit the static, detailed visual information provided by RGB imagery as effectively as traditional ANNs.

This pattern suggests a strategic placement of network types depending on the sensory data and environmental conditions. The use of SNNs with DVS data is particularly effective in bad weather due to their ability to robustly process dynamic visual changes. On the other hand, ANNs excel with RGB data in good weather, benefiting from the detailed color and texture information that is readily available. Such distinctions highlight the importance of selecting the appropriate neural network architecture tailored to specific sensor data and task requirements.

We have also noticed that for longer clips and more advanced tasks such as prediction SNN performance increases. For simple tasks like pedestrian crossing and good weather conditions performance of ANN and SNN are on pair with little advantage of ANN (as presented in tab. \ref{tab:clip-classification}).

Lastly, SNNs have shown great energy efficiency compared to ANNs. In some cases, the differences were orders of magnitude. Also, their energy usage seems not to be affected largely by the length of the analyzed sample. These findings are especially interesting from the perspective of AVs, as these systems must operate in energy-constrained environments. 

Overall, our findings advocate for a hybrid approach where the choice of neural network and sensory data type is aligned with specific operational environments to optimize pedestrian detection performance. This approach not only capitalizes on the strengths of each network type but also ensures adaptability across varying atmospheric conditions.

\section{Conclusions}

This study has successfully demonstrated the application of Spiking Neural Networks (SNNs) integrated with Dynamic Vision Sensors (DVS) for enhancing pedestrian detection in adverse weather conditions, which is crucial for advancing autonomous vehicle technologies. Our research focused on comparing the effectiveness of SNNs with traditional Convolutional Neural Networks (CNNs) across different sensory modalities and environmental settings.

Key findings from our experiments include:
\begin{itemize}
    \item \textbf{Enhanced Detection in Adverse Conditions:} SNNs, when combined with DVS, show superior performance in low-light and poor weather conditions, leveraging the high dynamic range and temporal resolution of DVS to detect subtle movements and changes in pixel intensity.
    \item \textbf{Energy Efficiency:} SNNs demonstrate significant reductions in energy consumption compared to traditional CNNs, making them ideal for real-time applications in autonomous driving where power efficiency and quick response times are paramount.
    \item \textbf{Challenges with RGB Data:} While SNNs excel with DVS data, their performance with standard RGB data remains a challenge, particularly under variable weather conditions where traditional CNNs tend to perform better.
    \item \textbf{Impact of Clip Length and Complexity:} Our findings suggest that SNNs potentially increase their performance as the complexity of the task increases, such as in longer clip lengths for pedestrian prediction, indicating a capacity for handling more detailed temporal sequences.
\end{itemize}

The research has also underscored the importance of choosing the correct neural network architecture and sensor modality tailored to specific operational environments, advocating for a hybrid approach where SNNs are deployed alongside traditional ANNs to optimize the strengths of each according to situational demands.

Future work will focus on addressing the challenges identified, particularly in enhancing the processing capabilities of SNNs with RGB data and further optimizing the integration of SNNs with DVS for broader applications in intelligent transportation systems. Additionally, exploring more advanced models and training techniques to further improve the robustness and accuracy of pedestrian detection under varying environmental conditions will be crucial.

\section*{Data and code availability}
The code for experiments is available at \href{https://github.com/szmazurek/snn\_dvs}{https://github.com/ szmazurek/snn\_dvs}. Instructions for downloading the dataset are also provided there.

\section*{Acknowledgments}
We gratefully acknowledge Poland's high-performance Infrastructure PLGrid ACK Cyfronet AGH for providing computer facilities and support within computational grant no. PLG/2023/016767. The publication was created within the project of the Minister of Science and Higher Education "Support for the activity of Centers of Excellence established in Poland under Horizon 2020" on the basis of the contract number MEiN/2023/DIR/3796. This publication is supported by the European Union’s Horizon 2020 research and innovation programme under grant agreement Sano No 857533. This publication is supported by Sano project carried out within the International Research Agendas programme of the Foundation for Polish Science, co-financed by the European Union under the European Regional Development Fund.

{\appendix[Labeling problems in JAAD dataset adoption]

Due to observed performance drops on the JAAD dataset, we performed additional error analysis by visualizing the test set samples for which the corresponding model assigned an incorrect label. We observed that irrespective of the task or used model, errors occurred on samples from specific clips. Some examples are visualized in Fig. \ref{fig:jaad-errors}. In those videos, the assigned labels from the original annotations can be considered as some form of "corner cases". For example, top right video, the man is directly crossing the path that the car is following. Therefore, based on other examples in the dataset, networks correctly predict that the crossing behavior is indeed happening or going to happen soon, yet this prediction is considered incorrect. This can be somewhat expected, as the JAAD dataset was focused mostly on the intention to cross.
In the simulation datasets proposed in this work, such inconsistencies are not present, thus allowing the experimenter to focus purely on the algorithm's effectiveness without considering the problems related to the data labeling.
\begin{figure*}[bt]
    \centering
    \includegraphics[scale=0.5]{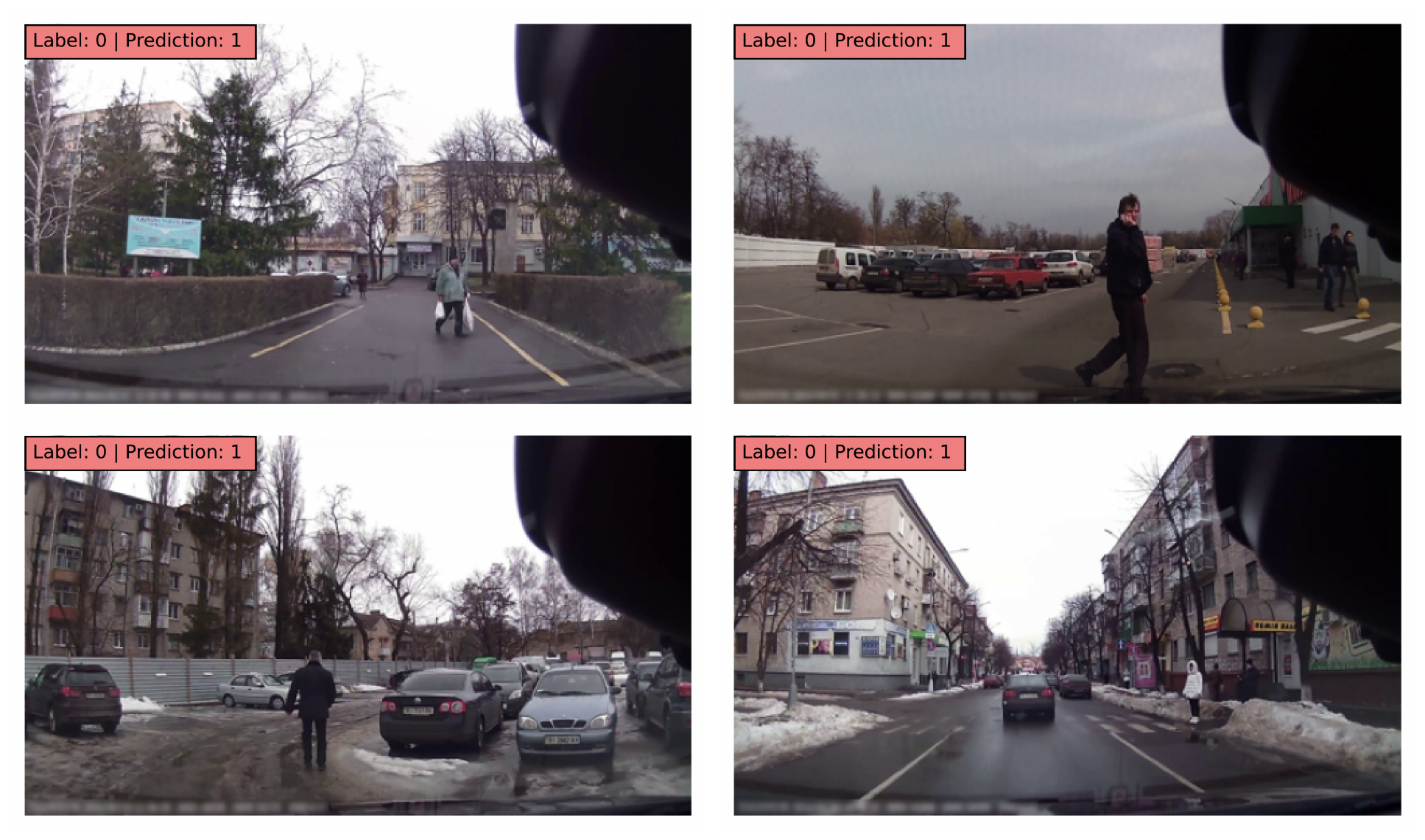}
    \caption{Examples of incorrectly classified frames evaluated on bad weather subset of JAAD dataset. The predictions were made by the SPS R18T model in a single-frame prediction task. In those examples it is visible that the labels assigned to the given frame do not precisely match the observed pedestrian behavior, therefore leading to the prediction being considered incorrect.}
    \label{fig:jaad-errors}
\end{figure*}
}

{\appendix[A note on DVS frames decimation]

As mentioned in the article, to lower the computational demands of the multiple experiments that were conducted, we decided to resize the images to 450x256 resolution from the original 1600x600 for RGB and 1542x587 for DVS modalities. To perform such an operation, one needs to leverage the interpolation procedure. As the interpolation methods use the values of a pixel neighborhood when computing the new values for the pixels in the resized image, it poses a potential problem when working with DVS data. This modality is based on very precise measurement of the light change magnitude at the given points, therefore potential approximations may degrade the quality. We explored this phenomenon and found that indeed, from the human observer perspective, the DVS images were significantly affected by the interpolation regardless of the method used. Examples of such transformations can be seen in Fig. \ref{fig:interpolation-noise}.
Despite these findings, we observed in initial trials that the networks trained on such data were still able to converge and reach high performances when using the nearest neighbor interpolation method. Therefore we proceeded with using such operations in the experiments shown in this paper. Other resizing techniques, such as seam carving or super-resolution networks, can potentially be used to solve this problem. We leave their evaluation for future research.

\begin{figure*}[tb]
    \centering
    \includegraphics[height=0.5\linewidth,width=1\linewidth]{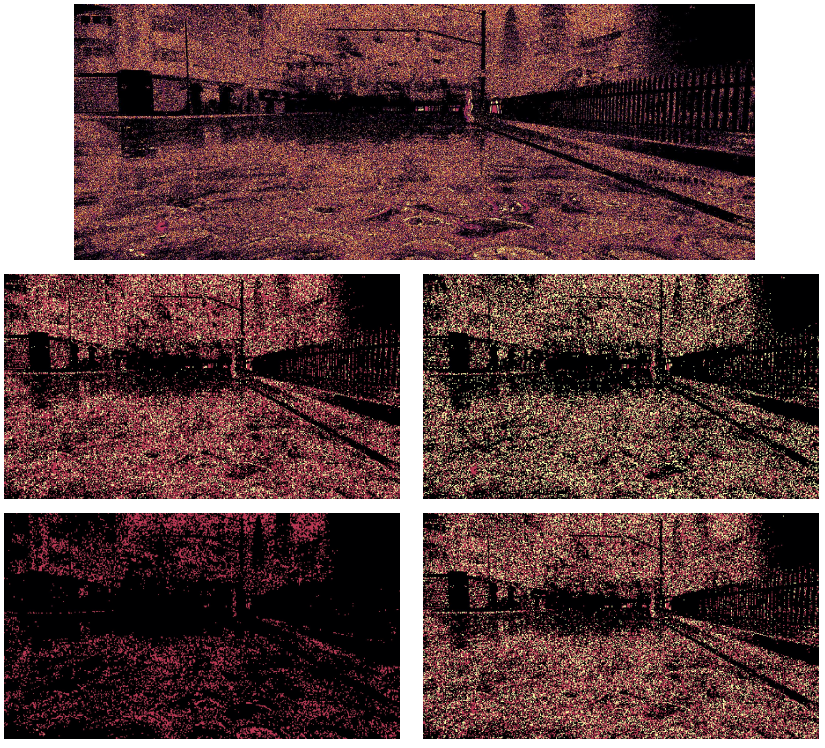}
    \caption{Comparison of original size DVS image with the pedestrian crossing the street in bad weather with the same frame after interpolation. The top image is the original frame, the grid below shows the interpolation results for different methods. Top left: bilinear, top right: nearest neighbor, bottom left: areal, bottom right: bicubic. Note that after the interpolation the image quality degrades significantly, making the visibility of the pedestrian much lower for the human observer.}
    \label{fig:interpolation-noise}
\end{figure*}
}

{
\appendix[Spiking Resnet18 modifications - ablation study]
We created our SNN by incorporating the findings of \cite{fangSEWResnet,duanTEBN} into the original ResNet18 architecture converted to the spiking variant. We constructed an ablation study to determine the effectiveness and influence of the given modification on the final model performance. We evaluated three models, namely baseline Spiking ResNet 18 (SP R18), Spiking SEW ResNet18 \cite{fangSEWResnet} (SPS R18) and the SPS R18 model with batch normalization layers replaced with Temporally Effective Batch Normalization \cite{duanTEBN}. We evaluated the networks in the single-frame detection task, where single input frame with associated label was shown to the network either once or repeated 10 times in a sample. The results can be seen in Tab. \ref{tab:ablation}.

With one input frame, SPS R18 shows remarkable improvement over the SP R18 network in all data cases, except for the DVS modality in the bad weather subset where the latter reached comparable AUROC. The addition of TEBN resulted in a similar performance as for SPS R18, surprisingly with slight favor for the one without the batch normalization modification. The exception was RGB data in the normal weather subset, where SPS R18T outperformed the SPS R18T. The first network reached an AUROC of 0.8541 and an F-score of 58.61\%, while the achieved one was 0.5686 and 34.2\% for both metrics respectively.

In the case of the input frame repeated 10 times in the input samples, SP R18 has shown large performance gains, even exceeding the enhanced versions in most cases. SPS R18 improved the performance with RGB data in the normal weather subset but lost it in the bad weather subset for the same modality compared to the one-frame-only approach. SPS R18T once again has shown stable performance across all modalities.

Examining those results, we finally chose the SPS R18T for further evaluation due to its stable, robust performance across most modalities. Even in the cases where it did not reach top results in a given setting, it remained close to the best-performing network. Finally, the ability of this network to perform nearly equally well when using samples consisting of only one frame with no repetitions was favorable, as this reduced latency directly increases the speed of computations and reduces energy usage in potential future applications.

\begin{table*}[]
\centering
\caption{Performance of the compared networks on different datasets as measured by test AUROC and F-score for single or repeated frame classification. F-score is denoted in percentages. Bold numbers indicate the best results for the given data subset. SPS R18  stands for the Spiking SEW ResNet18, TEBN stands for Temporal Effective Batch Normalization. Bold numbers indicate the best results for the given data subset and modality.}
\begin{NiceTabular}{|p{2cm}|p{1.cm}|p{1.cm}|p{1.cm}|p{1.cm}|p{1.cm}|p{1.cm}|p{1.cm}|p{1.cm}|}
    \hline
   \Block[draw]{2-1}{\diagbox{Network}{Subset}} &   \Block{1-4}{\makecell{Bad weather}} & & & & \Block{1-4}{\makecell{Normal weather}}  \\
   \cline{2-9}
   & \Block{1-2}{\makecell{DVS}} &  & \Block{1-2}{\makecell{RGB}} &  & \Block{1-2}{\makecell{DVS}} &  & \Block{1-2}{\makecell{RGB}} &  \\
   \hline
     \diagbox{}{} & AUROC & F-score & AUROC & F-score & AUROC & F-score  & AUROC & F-score  \\
     \hline
     \Block{1-9}{\makecell{Clip length: 1 frame}} & & &    \\
     \hline
     SP R18  & \textbf{0.9429} & 75.86 & 0.5594 & 29.94 & 0.8365 & 68.64 & 0.5 & 0 \\
     \hline 
    SPS R18  & \textbf{0.9478} & \textbf{79.97} & \textbf{0.7676} & \textbf{42.02} & \textbf{0.9301} & \textbf{75.28} & 0.5686 & 34.2  \\
    \hline 
    SPS R18T & \textbf{0.9467} & \textbf{79.16} & 0.741 & 39.38 & 0.8985 & 72.25 & \textbf{0.8541} & \textbf{58.61} \\
    \hline
     \Block{1-9}{\makecell{Frame repeats: 10 frames}} & & &    \\
     \hline
     SP R18  &  0.9637 &  78.57 & \textbf{0.737} & \textbf{42.58} & \textbf{0.9485} & \textbf{77.58} & \textbf{0.9116} & \textbf{68.97} \\
     \hline
     SPS R18  &  \textbf{0.9771} &  \textbf{83.08} & 0.5759 & 11.54 & \textbf{0.9421}  & 73.95 & 0.8735 & 63.4 \\
     \hline
     SPS R18T & 0.9636 & \textbf{83.61} & 0.7226 & 37.91 & 0.9288 & 74.22 & 0.8325 & 54.32 \\
     \hline
\end{NiceTabular}
\label{tab:ablation}

\end{table*}
}

\FloatBarrier

\bibliography{bibliography}


 




\vfill

\end{document}